# Neuromuscular Reinforcement Learning to Actuate Human Limbs through FES

Nat Wannawas[1], Ali Shafti[2], A.Aldo Faisal [1,2]

[1]Department of Bioengineering, Imperial College London, UK

[2]Department of Computing, Imperial College London, UK

a.faisal@imperial.ac.uk

***Abstract:*** Functional Electrical Stimulation (FES) is a technique to evoke muscle contraction through low-energy electrical signals. FES can animate paralysed limbs. Yet, an open challenge remains on how to apply FES to achieve desired movements. This challenge is accentuated by the complexities of human bodies and the non-stationarities of the muscles' responses. The former causes difficulties in performing inverse dynamics, and the latter causes control performance to degrade over extended periods of use. Here, we engage the challenge via a data-driven approach. Specifically, we learn to control FES through Reinforcement Learning (RL) which can automatically customise the stimulation for the patients. However, RL typically has Markovian assumptions while FES control systems are non-Markovian because of the non-stationarities. To deal with this problem, we use a recurrent neural network to create Markovian state representations. We cast FES controls into RL problems and train RL agents to control FES in different settings in both simulations and the real world. The results show that our RL controllers can maintain control performances over long periods and have better stimulation characteristics than PID controllers.

***Keywords:*** Functional Electrical Stimulation, Neuromuscular stimulation, FES, Reinforcement Learning, Automatic Control

## Introduction

Motor impairments and disabilities result in humans being unable or having limited ability in activating their muscles, to move their own limbs, heavily impacting their independence and quality of life. Functional Electrical Stimulation (FES) is an established and safe technique that can re-animate the paralysed limbs by stimulating the muscles with low-energy electrical signals. From robotic and control perspectives, FES enables the muscles to function as actuators through which the force is applied to manipulate the limbs. Yet, the capabilities in manipulating, for example, paralysed human arms through FES are relatively limited compared to what is capable in robot arms. These limited capabilities are substantially due to the complexities of human bodies as well as the non-linear and non-stationary responses of the muscles to the stimulation. These intertwined complexities post challenges in applying inverse dynamics, one of the core techniques in robotics, to FES systems. Additionally, most established control techniques in robotics were designed for stationary actuators such as motors. However, the actuators in FES systems, the muscles, have non-stationary responses to the control signal, the stimulation, mainly due to muscle fatigue.

Several control techniques are applied to control FES to manipulate the limbs; most of them are PID controller, Fuzzy Logic controller, and their variances [1]. These controllers perform reasonably well and can drive the error close to zero even when the muscles become fatigued. Yet, their performance in terms of response time and overshoots degrades in long sessions as they rely on the error signal to increase the stimulation in compensation to the fatigue. This problem is more pronounced in trajectory tacking tasks. The variances of sliding mode control show good performance in terms of fatigue compensation [2]. Yet, the implementations are slightly more complicated as they require the dynamics model of the systems. It is also worth noting that these aforementioned controllers themselves are only the intensity regulators; the control systems require additional muscle selection patterns that specify the muscles to stimulate given the states. Hence, the movements these controllers can evoke are limited to the well-understood motions such as single-joint flexion and cycling.

Advances in machine learning have resulted in more data-driven methods for control engineering, specifically, Reinforcement Learning (RL) presents great potential in control applications as substantiated by successes in controlling complex real-world systems such as robots [3]. RL can learn the whole stimulation policy which includes both intensity regulation and stimulation profile by itself, thereby potentially resulting in fully automatic control systems that can customise themselves to the users. RL in upper-limb FES control applications was studied in both simulation [4] and the real world [5]. These studies presented the potential of RL in FES applications. Yet, they were limited to brief point-to-point motions. RL in FES cycling was studied in [6]. This simulation-based study presents the ability of RL to learn to stimulate multiple muscles to control the cycling cadence at desired values against the advancing fatigue. However, those RL controllers estimate the levels of muscle fatigue through a fatigue model whose accuracies in the real world have yet to verify.

In this work, we present an RL system for controlling FES in long-duration trajectory tracking tasks which requires online muscle fatigue estimation to maintain the control performance. We approach this online estimation challenge by using a recurrent neural network, which can learn to summarise the history of state progression in its hidden states, to create Markovian state representations for the RL agent. We train our RL systems in different scenarios in a detailed biomechanics simulator that we enhanced with neurostimulation features. We then present the performance test of our controllers and the comparison with a conventional method. Finally, as a proof-of-concept, we validate our system in human volunteers.

# Methods

Here, we firstly present the architecture of our RL systems, followed by the general training setup to train the RL agents as FES controllers. Next, we present the setups of different FES control test scenarios which include the detailed musculoskeletal models and the real-world settings.

**RL Architecture:**

*Preliminary* − RL learns tasks through experiences obtained from the interactions with the environment described as follows. At each time step $t$, the agent observes an environment state vector $s_t$ and outputs an action vector $a_t=\pi_\theta(s_t)$, where $\pi_\theta$ is the policy parameterised by $\theta$. The environment affected by the action outputs immediate numerical reward $r_t$ and becomes in a new state $s_{t+1}$ which is observed by the agent in the next time step. The interaction experience is stored in a replay buffer as a tuple ($s_t$, $a_t$, $r_t$, $s_{t+1}$). The learning goal is to find a policy $\pi_\theta$ that maximises the expected return expressed as $R = \sum_{t=0}^{T} \gamma^t r_t$, where $\gamma \in [0,1)$ is a discount factor that emphasises the importance of immediate reward, and $T$ is the episode length.

*Architecture of our RL system* − Our RL system is divided into two modules: policy and state representation modules.
*1) The policy module* is responsible for learning the control policy. The learning mechanism varies across different RL algorithms. Here, we use Soft Actor-Critic (SAC) with double Q-networks [3], one of the state-of-the-art algorithms with robust performance across different environments. The policy is parameterised by a neural network (NN) with a Sigmoid activation function at the output layer to squash the action, which is the normalised stimulation intensity, within [0,1].
*2) The state representation module* is responsible for converting the observation into state representations. We use a recurrent neural network (RNN) named GRU [7], an easy-to-train RNN with state-of-the-art performances. We use GRU with 20 hidden units, selected based on empirical performances. A fully-connected layer is added after hidden units to map the hidden states to the observable states during the training. The GRU takes concatenation vectors of the state and action as inputs to update its hidden states whose values are used as the state representation through which the RL agent observes the environment.

**RL Training:** The training is divided into two phases: the interaction and the learning phases: These phases occur alternately in an episodic manner described as follows.

*The interaction phase* of an episode starts with random initial states and a random trajectory to track. The hidden states of the GRU are initialised by feeding the initial states and zero action vector to the GRU. The RL agent then observes a state vector comprising the values of the GRU's hidden states, which are the state representation, and the control target of the time step $s_{tar,t}$. After that, the interaction follows standard RL procedure, except that the agent observes the environment's states via the state representation which is updated every time step by the GRU (Fig. 1a).

*The learning phase* starts with the learning of the state representation module which is trained to predict $s_{t+1}$ from the input ($s_t$, $a_t$) using the time-series of state progression collected in each episode (Fig. 1b). After that, the experiences tuples in the replay buffer are fed to the GRU to convert the raw state observations into the state representations. The converted experience tuples are stored in a temporary replay buffer. Additionally, we augment the temporary buffer with hindsight experiences [8], which is carried out by replicating the real tuples and re-compute the reward by using $s_{tar,t} = s_{t+1}$. This process simply generates fake experiences of the agent reaching the targets. After that, the agent samples the tuples from the temporary buffer to update the policy as in the standard RL procedure.

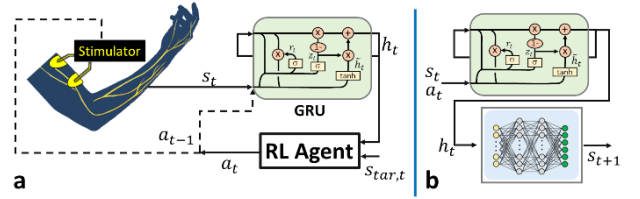

Figure 1: (a) Interactions between the components in the system during real-time, online control. (b) A fully-connected layer added after the recurrent layer to map the hidden states to the observable states in the learning phase.

**Conventional controller:**
We compare our RL controller to a conventional model-free controller for which a standard PID controller is chosen as it is one of the wildly used FES controllers [1]. PID controllers compute the stimulation intensity based on the errors and the gain parameters $K_P$, $K_I$, $K_D$; the higher gain, the more sensitive the controller is. Here, the gains are tuned using a genetic algorithm named CMA-ES [9].

**Simulation-based experiment setups:** The simulation experiments to investigate the performance of RL and PID controllers are carried out in 3 scenarios: *vertical arm motion* (Fig. 2a), *horizontal arm motion* (Fig. 2c), and *cycling motion* (Fig. 2d). The models are created in OpenSim, an open-source biomechanical simulation software, into which we add a muscle fatigue feature that causes the decline in muscles' force production capability based on the past stimulation (see [6] for the details). The models take normalised stimulation intensity as control inputs. The episode length of all scenarios is equivalent to 90 seconds in the real world. Each scenario has a different model and task to the others, described as follows.

The *Arm vertical motion* scenario has an arm model with two joints, shoulder (locked) and elbow, which moves on a sagittal plane (Fig. 2a). The model has 6 muscles: 3 from each of the Biceps and Triceps groups. The task here is to control the elbow angle through a single channel the stimulation $a_t \in [0,1]$ which is distributed equally across Biceps muscles. The observable state vector $s_t \in \mathbb{R}^3$ comprises the angle $\theta_t$, angular velocity $\dot{\theta}_t$, and target angle $\theta_{tar,t}$ of the elbow. The immediate reward $r_t$ is computed as $r_t = \sqrt{(\theta_t - \theta_{tar,t})^2} - a_t^2$.

The *Arm horizontal motion* scenario has similar settings to the vertical scenario, except that the arm, which is supported by rollers for gravity compensation, moves on a horizontal plane (Fig. 2c). Additionally, the stimulation has two channels ($a_t \in [0,1]^2$), which are for the Biceps and Triceps groups. The observable states and the control task are similar to those of the vertical scenario. The immediate reward is computed as $r_t = \sqrt{(\theta_t - \theta_{tar,t})^2} - \frac{\sum_{i=1}^{2} a_{t,i}}{2}$.

The *Cycling motion scenario* has a lower limb model with a cycling crankset that rotates without friction. (Fig. 2d). The pedals are attached to the feet, allowing the force transfer. The model has 18 muscles in total; 6 muscles which are rectus femoris, gluteus maximus, and hamstrings on both legs are stimulated ($a_t \in [0,1]^6$). The observation vector $s_t \in \mathbb{R}^3$ comprises the angle $\theta_t$, angular velocity (cycling cadence) $\dot{\theta}_t$, and target cadence $\dot{\theta}_{tar,t}$ of the crank. The task is to control the cadence at target values. The reward is computed as $r_t = \sqrt{(\dot{\theta}_t - \dot{\theta}_{tar,t})^2} - \frac{\sum_{i=1}^{6} a_{t,i}}{6}$.

**Real world experiment setups:**
As a proof-of-concept, we investigate and compare the performance of the RL controllers in the real world. This setup replicates the *arm vertical motion* scenario on healthy human volunteers. The stimulation pulses, which is generated by Rehastim 1 (HASOMED GmbH, Magdeburg, Germany) stimulator, are applied to Biceps muscles via a pair of surface electrodes (Fig. 2b). The elbow angles are recorded using an inertia measurement unit (IMU). The performance tests of RL and PID controllers were carried out on different days to minimise the effect of residual fatigue.

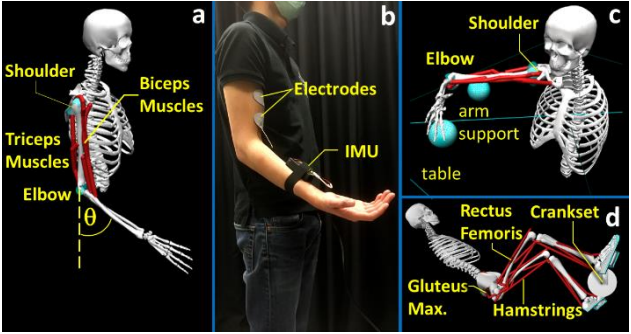

Figure 2: Our 3 test scenarios for FES control: (a) arm vertical motion in simulation and (b) human volunteers, (c) arm horizontal motion and (d) cycling motion.

## Results

**RL training results:**
The training is monitored by the average episodic error. Fig. 3 shows the episodic errors of all simulation scenarios plotted against the training episodes. We conducted 5 training repetitions to investigate the robustness of the training, which is a stochastic process. In vertical and horizontal arm motion scenarios, the RL controllers have the average final performance with an error around 5° (Fig. 3 a,b). The final performance in the cycling motion has an error of approximate 10 RPM (Fig. 3c).

For the real-world scenario, the RL controller is pre-trained using the vertical arm model whose muscle forces are optimised to match the real motion. After that, the pre-trained RL is trained further for 5 90-second-long episodes in the real world. Note that all RL controllers are trained to track random trajectories during the training.

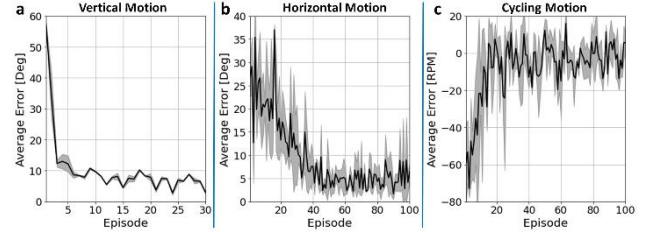

Figure 3: Average tracking error in (a,b) degrees for arm motion cases and in (c) RPM for cycling cases, against training episodes. The solid line and the shaded area show the mean and the standard deviation of 5 training runs.

**Tracking performance test:**
*Arm vertical motion* – We compare RL and PID controllers' performance in 2 three-minute-long tracking tasks. The first task is to track an arbitrary angle trajectory that includes ramping. Fig. 4 a and c show the performance of PID and RL controllers, respectively. Both controllers can track the trajectory. The PID controller has insignificantly better performance in the root mean square error (RMSE) measure. However, the RL controller produces fewer overshoots and better prepares for the drop of the arm as highlighted by blue circles. The second task is to track a trajectory that periodically sets the angle at 70 and 20 degrees. These two angles are selected to be in regions that are relatively difficult to maintain. Again, both controllers can track this trajectory, as seen in Fig. 4 b and d. The overshoots of the PID controller increase as the muscle becomes more fatigued, evidenced by the downward trajectory of the brown lines which show the percentage of maximum force that the muscles can produce – an established fatigue effect, which has to be compensated by increases in stimulation intensity (green lines).

*Arm horizontal motion* – The task is to control the elbow angle of the arm moving on the horizontal plane by stimulating the Biceps and Triceps muscles to track a 6-minute-long trajectory (Fig. 4 e and f). The RL controller can track the trajectory with slightly better RMSE than the PID controller which has growing overshoots as the muscles become fatigued. Additionally, the RL controller learns to apply co-contraction to prevent the overshoots and keep the angle still at the targets.

*Cycling motion* – The task here is to control the cycling cadence by stimulating 6 leg muscles. Fig. 4 g and h show the tracking behaviours of PID and RL controllers, respectively. The behaviours are quite similar. RL controller has slightly better RMSE measures and can reach the new target cadences slightly earlier than the PID controller with the muscle selection pattern and gain parameters adopted from [10].

*Arm vertical motion in the real world* – We validate our RL system experimentally on human volunteers. Both RL and PID controllers are tested to track the same trajectories as in the vertical motion simulation scenario. Fig. 4 e-f show that both PID and RL controllers perform poorer in the real world. In the first trajectory, the PID controller (Fig. 4i) produces good tracking in the first half but unstable control in the second half of the trajectory, especially along the ramping section. Our RL controller produces more consistent control along the same trajectory (Fig. 4j). In the second case, the PID controller (Fig. 4k) produces unstable control at the 70° targets. However, it still has a slightly better RMSE. The RL controller, again, has more consistent control (Fig. 4l).

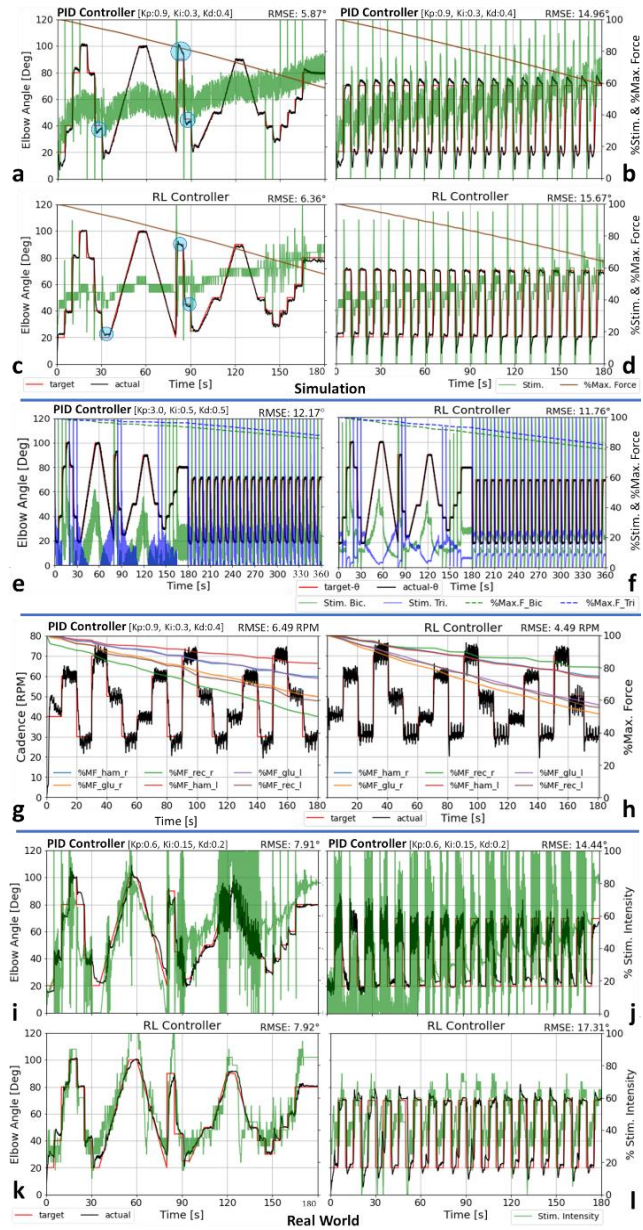

Figure 4: Simulation results of vertical arm motion task of (a,b) PID and (c,d) RL controllers. Experimental results of (i,j) PID and (k,l) RL in the same task. Simulation results of horizontal arm motion of (e) PID and (f) RL controllers. Results of cycling motion of (g) PID and (h) RL controllers. %Max. force in (a-f), shorten to %MF in (g-h), show the decline of maximum muscle force due to fatigue.

## Conclusion and Discussion

We present the application of RL in FES controls together with a method that enables RL to work in these non-Markovian systems. We test our RL controllers in long-duration trajectory tracking scenarios, including a real-world setting. The results show that our RL controllers can learn in these non-Markovian systems as they can appropriately increase the stimulation to compensate for the fatigue. In comparison to PID controllers, our RL controllers have yet to significantly outperform those PID controllers by RMSE measure. Yet, there are a few points that are worth mentioning. Firstly, RL controllers apply better stimulation as evident in the green lines in Fig. 4 i-l that show lower fluctuation in stimulation intensity in RL cases with higher tracking stability. Secondly, RL learns end-to-end stimulation policies, including muscle selection patterns. This means RL has the potential to evoke any possible motions; this work is a stepping stone to the advancement.

## Acknowledgement

Funding by a Royal Thai Govt. Scholarship to NW and a UKRI Turing AI Fellowship (EP/V025449/1) to AAF.

## References

[1] N. Dunkelberger et al., "A review of methods for achieving upper limb movement following spinal cord injury through hybrid muscle stimulation and robotic assistance," *Experimental Neurology,* 2020.

[2] J. Saso et al., "Sliding Mode Closed-Loop Control of FES: Controlling the Shank Movement," *IEEE Trans. Biomed. Eng.,* 2004.

[3] T. Haarnoja et al., "Soft Actor-Critic Algorithms and Applications," *arXiv [cs.LG],* 2019.

[4] P. S. Thomas et al., "Creating a Reinforcement Learning Controller for FES of a Human Arm," *Yale Workshop Adapt. Learn Syst.,* 2008.

[5] F. D. Davide et al., "Does Reinforcement Learning outperform PID in the control of FES-induced elbow flex-extension?," *IEEE Intl. Workshop Med. Mea. and Appl. (MEMEA),* 2018.

[6] N. Wannawas et al., "Neuromechanicsbased Deep Reinforcement Learning of Neurostimulation Control in FES Cycling," in *Intl. Conf. Neural Engineering (NER)*, 2021.

[7] J. Chung et al., "Empirical Evaluation of Gated Recurrent Neural Networks on Sequence Modeling," in *Proc. Conf. Neu. Info. Pro. Sys. (NIPS)*, 2014.

[8] M. Andrychowicz et al., "Hindsight Experience Replay," in *Conf. Neural Info. Pro. Syst.*, 2017.

[9] N. Hansen, "The CMA Evolution Strategy: A Tutorial," *arXiv [cs.LG],* 2016.

[10] A. Sousa et al., "A Study on Control Strategies for FES Cycling Using a Detailed Musculoskeletal Model," *IFAC-PapersOnLine,* 2016.